\documentclass[conference]{IEEEtran}
\IEEEoverridecommandlockouts

\usepackage{cite}
\usepackage{amsmath,amssymb,amsfonts}
\usepackage{algorithmic}
\usepackage{graphicx}
\usepackage{textcomp}
\usepackage{xcolor}
\def\BibTeX{{\rm B\kern-.05em{\sc i\kern-.025em b}\kern-.08em
    T\kern-.1667em\lower.7ex\hbox{E}\kern-.125emX}}

\usepackage{hyperref}
\usepackage{cleveref}
\usepackage{booktabs}
\usepackage{makecell}

\begin{document}

\title{{\footnotesize \textit{To appear in Proceedings of the 2025 IEEE International Conference on Data Science and Advanced Analytics (DSAA 2025)}\\[1.0em]} HyperKD: Distilling Cross-Spectral Knowledge in Masked Autoencoders via Inverse Domain Shift with Spatial-Aware Masking and Specialized Loss}

\author{
\IEEEauthorblockN{Abdul Matin}
\IEEEauthorblockA{\textit{Colorado State University} \\
Fort Collins, Colorado, USA \\
amatin@colostate.edu}
\and
\IEEEauthorblockN{Tanjim Bin Faruk}
\IEEEauthorblockA{\textit{Colorado State University} \\
Fort Collins, Colorado, USA \\
tanjim@colostate.edu}
\and
\IEEEauthorblockN{Shrideep Pallickara}
\IEEEauthorblockA{\textit{Colorado State University} \\
Fort Collins, Colorado, USA \\
shrideep@colostate.edu}
\and
\IEEEauthorblockN{Sangmi Lee Pallickara}
\IEEEauthorblockA{\textit{Colorado State University} \\
Fort Collins, Colorado, USA \\
sangmi@colostate.edu}
}

\maketitle
\begin{abstract}

The proliferation of foundation models, pretrained on large-scale unlabeled datasets, has emerged as an effective approach in creating adaptable and reusable architectures that can be leveraged for various downstream tasks using satellite observations. However, their direct application to hyperspectral remote sensing remains challenging due to inherent spectral disparities and the scarcity of available observations. In this work, we present HyperKD, a novel knowledge distillation framework that enables transferring learned representations from a teacher model into a student model for effective development of a foundation model on hyperspectral images. Unlike typical knowledge distillation frameworks, which use a complex teacher to guide a simpler student, HyperKD enables an inverse form of knowledge transfer across different types of spectral data, guided by a simpler teacher model. Building upon a Masked Autoencoder(MAE) with a Vision Transformer (ViT) backbone, HyperKD distills knowledge from Prithvi (a ViT-based MAE geospatial foundation model trained on lower-dimensional multispectral data) into a student tailored for EnMAP hyperspectral imagery. HyperKD addresses the inverse domain adaptation problem with spectral gaps by introducing a feature-based strategy that includes spectral range-based channel alignment, spatial feature-guided masking, and an enhanced loss function tailored for hyperspectral images. HyperKD bridges the substantial spectral domain gap, enabling the effective use of pretrained foundation models for geospatial applications. Extensive experiments show that HyperKD significantly improves representation learning in MAEs, leading to enhanced reconstruction fidelity and more robust performance on downstream tasks such as land cover classification, crop type identification, and soil organic carbon prediction, underpinning the potential of knowledge distillation frameworks in remote sensing analytics with hyperspectral imagery.

\end{abstract}





\begin{center}

© 2025 IEEE. Personal use of this material is permitted. Permission from IEEE must be obtained for all other uses, in any current or future media, including reprinting/republishing this material for advertising or promotional purposes, creating new collective works, for resale or redistribution to servers or lists, or reuse of any copyrighted component of this work in other works.

\end{center}

\section{Introduction}
Hyperspectral satellites capture images of the Earth using hyperspectral imaging technology, with sensors mounted on satellites. Compared to traditional multispectral satellite images that utilize RGB bands or multispectral bands with 3–15 wavelengths (e.g., LandSAT, Sentinel-2), these images consist of pixels that represent observations across hundreds of narrow wavelength ranges, creating a detailed spectral fingerprint of an object's physical properties and chemical composition. As a result, hyperspectral satellite images enable advanced analyses such as identifying tree and plant species, monitoring air quality (both particulate and gaseous), assessing water quality (e.g., detecting algal blooms), and observing tree health with greater accuracy. Beyond its fundamental ability to provide extensive information, recent advances in miniaturizing system materials, improving data processing software, and reducing satellite launch costs have driven several world governments and leading private companies to launch \cite{storch2023enmap,loizzo2018prisma} or invest \cite{planet2023} in hyperspectral satellites over the past few years.

To leverage spectral information collected from satellite-based remote sensing, \textit{foundation models} have emerged as effective large-scale, pretrained architectures capable of capturing versatile representations for diverse sensing domains. This makes them well-suited for a broad range of downstream applications. Recent advances in foundation models have led to transformative breakthroughs in both computer vision and natural language processing, and similar approaches are increasingly being applied to the analysis of remote sensing data \cite{hong2024spectralgpt,liu2024remoteclip,sun2022ringmo}. For example, foundation models trained on multispectral satellite images, such as Prithvi \cite{jakubik2023foundation}, SatMAE \cite{cong2022satmae}, and ScaleMAE \cite{reed2023scalemaescaleawaremaskedautoencoder} have shown strong performance in analyzing geospatial data over large areas with high accuracy and robustness. 

However, developing a foundation model with hyperspectral images introduces critical challenges. High resolution spectral information in hyperspectral satellite images significantly increases the dimensionality of the data, introducing substantial computational challenges in both analysis and interpretation. In particular, modeling techniques with high computational demands (e.g., deep learning models) face major obstacles when adapting to the complexity of hyperspectral data.

Furthermore, hyperspectral satellite images with frequent cloud obscuration and extremely sparse revisit frequencies present significantly limited data availability. For example, EnMAP, one of the most recent hyperspectral satellites, has a revisit frequency of 27 days, compared to the 2-3 days visit cycle of the widely used NASA’s Harmonized Landsat Sentinel-2 \cite{claverie2018harmonized}. These limitation poses challenges for model performance, particularly in terms of generalization, which is critical for ensuring effectiveness and accuracy of pretrained foundation models in downstream applications.

In this study, we explore methodologies for pretraining a foundation model on hyperspectral satellite imagery, aiming to generate highly representative outputs that enhance the performance of downstream models. To address the aforementioned challenges posed by the characteristics of hyperspectral data, we propose a novel knowledge distillation framework, \textit{HyperKD}, that leverages knowledge from an existing pretrained foundation model that is originally trained on multispectral satellite imagery with lower spectral resolution but significantly higher revisit frequency and broader spatial coverage.

Knowledge Distillation (KD) offers an effective solution by transferring learned representations from a large \textit{teacher} model into a smaller, more targeted \textit{student} model. This enables the student model to inherit crucial high-level features while efficiently adapting to new tasks, datasets, and operational constraints—delivering the power of a foundation model without the heavy computational cost. Two key challenges often arise in the domain of KD: discrepancies in model architectures, referred to as the \textit{model gap}, and disparities in input data distributions, known as the \textit{domain gap} \cite{gou2021knowledge}. Most research focuses on narrowing the model gap — distilling knowledge from a high-capacity teacher network into a smaller student model. But in hyperspectral remote sensing, the domain gap is just as critical, requiring models to bridge differences in spectral resolution and data distributions. Yet, bridging this domain gap for hyperspectral imaging remains relatively unexplored.

With HyperKD, we reverse the conventional paradigm and propose a novel KD domain adaptation methodology that transfers knowledge from a \textit{smaller teacher model} with lower-dimensional inputs to a \textit{more complex student model} with higher-dimensional inputs. This approach captures high-level knowledge distilled at the level of spectral groups while also modeling interdependencies between spectral measurements at different geospatial locations. 

HyperKD leverages Prithvi-100M \cite{jakubik2023foundation}, a state-of-the-art foundation model trained on lower-dimensional multispectral data  (HLS \cite{claverie2018harmonized}, 6 bands, available every 2-3 days) across the entire CONUS, as the teacher model. A student model is then designed using EnMAP hyperspectral satellite data, which provides 218 bands but with a much lower revisit frequency (every 27 days), focusing on the California region to demonstrate its effectiveness. 

We integrate a spatial feature-guided masking strategy, which directs the student model to distill insights from the teacher prioritizing the reconstruction of the most challenging regions in input images. The feature score is estimated for each small region or patch using a Gabor Filter or Wavelet Transform to select the salient patches for masking. This approach is augmented with a feature-structure-aware reconstruction loss to ensure the retention of structural integrity and enhance the learning process by incorporating the structural similarity index metric (SSIM). Our key contributions can be summarized as follows:
\begin{itemize}
    \item \textbf{A Novel Cross Domain KD Architecture for MAE on HSI}: HyperKD bridges the spectral gap, adapting the inverse domain shift within a masked autoencoder framework for HSI.
    
    \item \textbf{Spatial Features Driven Masking Strategy}: HyperKD introduces an advanced masking strategy that leverages feature selection methods to distill insights from the foundational teacher model, specifically addressing the challenge of spatial variability while enhancing training efficacy.
    
    \item \textbf{Specialized Loss Function}: We incorporate a custom loss formulation that prioritizes feature importance, enhancing the fidelity of transferred representations and resulting in improved performance for hyperspectral applications. 
\end{itemize}

\section{Related Work}

\subsection{Foundation Models for Geospatial Images}
Foundation models have transformed remote sensing by enabling large-scale, pre-trained architectures capable of learning versatile representations from geospatial data. Models like \textbf{Prithvi} \cite{jakubik2023foundation} and \textbf{SatMAE} \cite{cong2022satmae}
have demonstrated the effectiveness of foundation models in capturing rich spatial and spectral features from satellite imagery. These models leverage massive datasets, such as Harmonized Landsat Sentinel-2 (HLS) and Sentinel-2, to achieve state-of-the-art performance in tasks like land cover classification, change detection, and disaster monitoring. For instance, Prithvi, a transformer-based foundation model trained on continental-scale multispectral data, has shown remarkable generalization capabilities across diverse geospatial tasks. Similarly, \textbf{ScaleMAE} \cite{reed2023scalemaescaleawaremaskedautoencoder} has introduced scalable self-supervised learning for remote sensing, further advancing the applicability of foundation models in this domain. 

\subsection{Knowledge Distillation Approach}
Knowledge distillation (KD) has emerged as a powerful technique for transferring knowledge from large, pre-trained models (teacher) to smaller, task-specific models (student) in various ways \cite{gou2021knowledge}. Originally proposed by Hinton \cite{hinton2015distillingknowledgeneuralnetwork}, KD has been widely adopted in computer vision and natural language processing. In remote sensing, KD has been applied to bridge the gap between large foundation models and resource-constrained applications. Recent advancements in KD, such as feature-based distillation \cite{park2023knowledge} and attention-based distillation \cite{shamsolmoali2023efficient}, have further improved the transfer of high-level features from teacher to student models. 

\subsection{Knowledge Distillation for Hyperspectral Images}
While KD has been extensively studied for multispectral satellite imagery, its application to HSI remained relatively underexplored until recently. Recent works \cite{9360315, chi2022novel,sun2023siamohot,yang2023random} have explored self-supervised learning (SSL) and adaptive distillation to effectively leverage unlabeled data. For instance, \textbf{SSKD} introduces adaptive soft labels by considering spatial and spectral similarities between labeled and unlabeled samples \cite{9360315}. Another advancement integrates adaptive knowledge distillation with 3-D transformations to better exploit spatial-spectral features in HSI cubes \cite{chi2022novel}. 

These studies highlight the potential of KD to enable efficient and accurate analysis of hyperspectral data.
In our work, we focus on cross-domain KD from a pre-trained foundation model, leveraging its capabilities for efficient HSI based downstream applications. 

\subsection{Use Cases of KD-Based Models in Remote Sensing}
KD-based models have been successfully applied to a wide range of remote sensing tasks, demonstrating their versatility and effectiveness. In land cover classification, Yue \textit{et al}. \cite{yue2021self} used KD to solve scarcity of labeled samples in HSI, proposing a self-supervised learning (SSL) method with adaptive distillation to train the deep neural network.
 Liu \textit{et al}. \cite{liu2024multimodal} proposed a multi-modal distillation framework to improve the robustness of student models in identifying land designed \textbf{STRM-KD} \cite{li2025strm} using topological relation matching KD for leaf disease recognition. These use cases underscore the potential of KD to address the computational and domain-specific challenges in remote sensing.

\section{Dataset and Study Area}


We utilize hyperspectral data from the Environmental Mapping and Analysis Program (EnMAP) \cite{storch2023enmap}, which originally spans 224 spectral bands in the 420–2450 nm range with a spatial resolution of 30 meters. 

For our study, initially, we collected cloud-free data between 2022 and 2024 from the state of California, an area characterized by diverse land-cover types, including expansive urban regions, dense forests, and varied agricultural land uses. Later we extended it to Colorado and Kansas to evaluate the model's robustness and generalization.


Following the removal of 6 noisy bands and preliminary data cleaning, our analyses focus on 218 bands. Additionally, we resampled the EnMAP scenes into smaller tiles of size $224 \times 224$ pixels to ensure compatibility with our foundational model (FM), Prithvi, which was pre-trained on Harmonized Landsat and Sentinel-2 (HLS) imagery (more details are available in Section~\ref{sec:hls_elmap_background}).

We partitioned the tiled data into four subsets: (i) a training set of 5000 tiles, (ii) a validation set of 500 tiles, (iii) a test set (Test Dataset 1) of 2000 tiles, and (iv) a separate test set (Test Dataset 2) of 2000 tiles with no spatial overlap to ensure robust evaluation. We considered another combined dataset across California, Colorado, and Kansas regions with 4625 tiles for training, and 1320 tiles for testing ( Test Dataset 3). Due to computational constraints, certain ablation experiments were conducted on reduced subsets of these datasets.

\section{Methodology}
\begin{figure*}[!htbp]
  \centering
    \includegraphics[height=8cm, width=0.85\textwidth, alt={Proposed architecture diagram}]{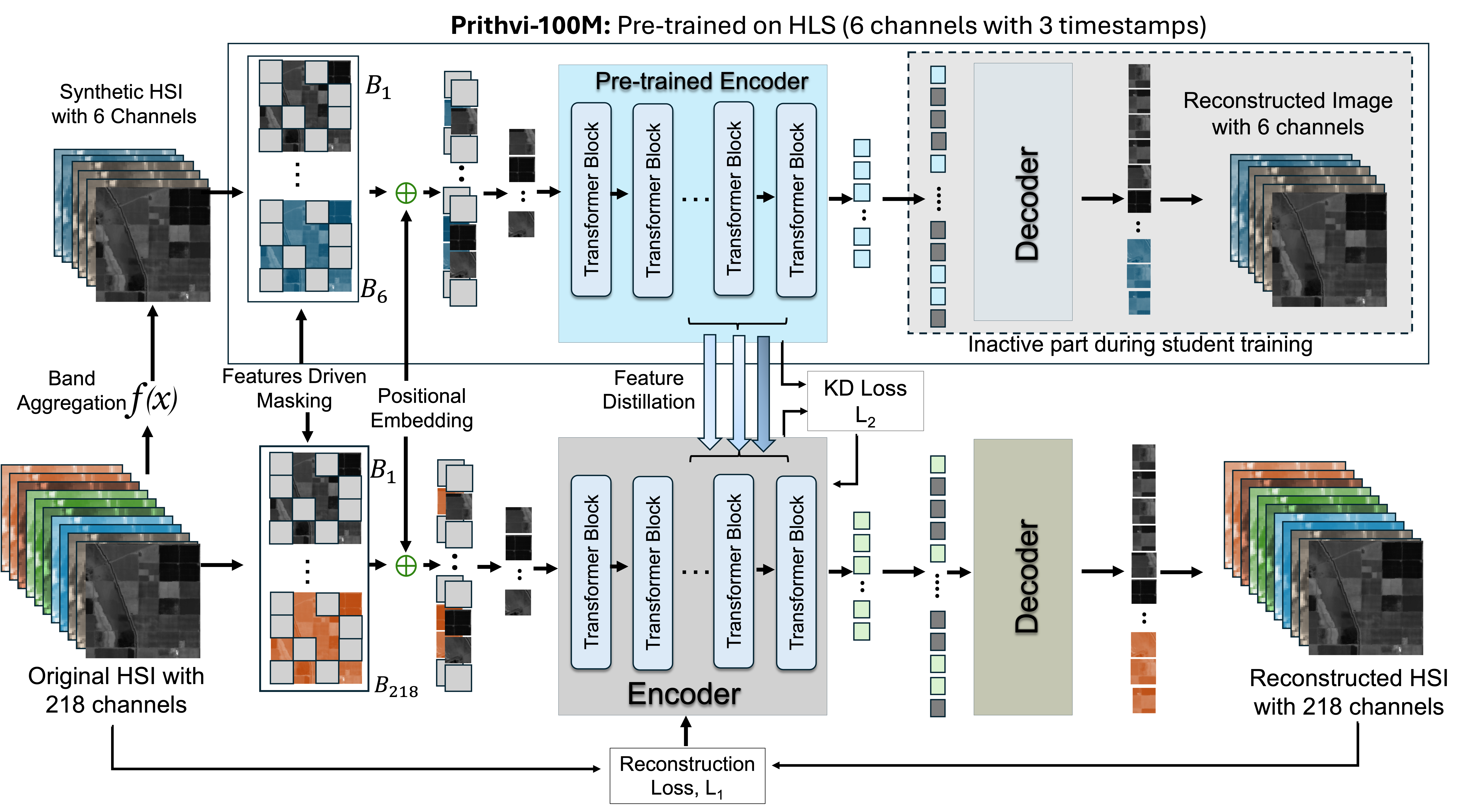}
  \caption{Architecture of HyperKD: A novel cross-domain knowledge distillation framework for HSI adapting in-middle layer insights from Prithvi's multispectral knowledge by employing guided masking and specialized loss function.}
  \label{fig:prop_architecture}
  \vskip -0.15in
\end{figure*}


\autoref{fig:prop_architecture} illustrates our methodology for transferring knowledge from a foundation geospatial autoencoder—pre-trained on 6-band HLS multispectral data—to a student model that (another masked autoencoder framework) processes 218-band EnMAP hyperspectral imagery, tackling the challenge of cross-channel knowledge transfer. Our methodology includes: (1) a channel-extended knowledge distillation strategy that improves teacher-student alignment across a broader spectral range, (2) a spatial feature-guided patch-scoring mechanism (employing Gabor filters and wavelets) to selectively mask high-confidence patches, and (3) a custom feature distillation and reconstruction loss tailored specifically for hyperspectral data.



\subsection{Bridging the Spectral Gap via Cross-Domain Knowledge Distillation}

\subsubsection{Background: Spectral Properties of HLS and ENMAP Data}
\label{sec:hls_elmap_background}

The Harmonized Landsat Sentinel-2 (HLS) project produces integrated surface reflectance data by merging observations from the Operational Land Imager (OLI) on Landsat satellites and the Multi-Spectral Instrument (MSI) on Sentinel-2 satellites. Although HLS imagery encompasses multiple spectral bands, the foundational model (FM) used in this study—Prithvi, developed by NASA and IBM—relies on only 6 bands for pretraining. These bands cover the visible to shortwave infrared (SWIR) range, covering approximately 450 nm to 2200 nm. In contrast, the EnMAP dataset provides a substantially higher spectral resolution, delivering 224 bands between 420 nm and 2450 nm. 

\subsubsection{Temporal Coverage Constraints}

The FM model is pre-trained on HLS data captured at three distinct timestamps, \textit{i.e.}, for each spatial location, three images are used to account for short-term temporal variations (every 2–3 days). By contrast, EnMAP has a much longer revisit interval (about 27 days), which complicates any attempt to align three consecutive EnMAP captures with the FM's more frequent HLS acquisitions. 
Moreover, these EnMAP acquisitions would need to align closely with the HLS timestamps to avoid discrepancies in atmospheric conditions, vegetation growth, or seasonal change.
Consequently, we restrict EnMAP observations to a single timestamp for each location, selecting the date that most closely aligns with any one of the teacher model's HLS timestamps.
\subsubsection{Aligning HLS and EnMAP Spectral Domains}
\label{sec:hls_enmap_alignment}

A key obstacle in distilling knowledge from a teacher model pre-trained on HLS data to a student model fine-tuned on EnMAP data is the spectral misalignment between the two datasets. While HLS provides only 6 spectral bands, EnMAP offers 218 bands with more finely resolved wavelength intervals. Furthermore, the spectral bands of HLS and EnMAP are not inherently aligned in terms of their wavelength ranges. Despite the spectral gaps, these image types are closely related due to their shared nature of capturing spectral attributes over a continuous spatial extent. This relationship has been widely utilized in spectral modeling \cite{zhang2022survey} such as the Lambertian assumption \cite{koppal2021lambertian}, which defines the relationship between hyperspectral and RGB images. Accordingly, without properly aligning EnMAP's spectral profiles to HLS's band definitions, the teacher model would receive inputs outside its expected feature space, undermining the effectiveness of cross-domain knowledge transfer. To address this, we employ a spectral matching procedure that reorders and aggregates EnMAP's hyperspectral bands to form synthetic bands that closely match the spectral coverage of the HLS bands used by the FM. 



Specifically, for each of the 6 spectral bands in HLS, we identify the EnMAP bands whose wavelength ranges overlap with the corresponding HLS band's spectral coverage. We then aggregate these overlapping EnMAP bands into synthetic bands that simulate the spectral characteristics of HLS.

Formally, we define HLS and EnMAP bands as follows:

\begin{equation}
    B_{\text{HLS}} = \{b_1, b_2, \ldots, b_6\}
\end{equation}

where each band \( b_i \) has a spectral range \( [\lambda_i^{\min}, \lambda_i^{\max}] \) and

\begin{equation}
       B_{\text{EnMAP}} = \{e_1, e_2, \ldots, e_{218}\}
\end{equation}

where each band \( e_j \) has a spectral range \( [\lambda_j^{\min}, \lambda_j^{\max}] \).
 


For each HLS band $b_i$, we define the subset $E_i \subseteq B_{\text{EnMAP}}$ that captures all EnMAP bands overlapping with the spectral interval of $b_i$:

\begin{equation}
E_i = \{ e_j \in B_{\text{EnMAP}} \mid \lambda_j^{\min} \geq \lambda_i^{\min} \ \text{and} \ \lambda_j^{\max} \leq \lambda_i^{\max} \}.
\end{equation}

This step reorders and creates subsets of EnMAP's hyperspectral coverage so that it directly corresponds to each HLS band's (e.g. RGB, NIR, SWIR) range. 

We then aggregate the EnMAP bands in \( E_i \) to create a synthetic band \( \hat{b}_i \) corresponding to the HLS band \( b_i \) by taking the average: 


\begin{equation}
\hat{b}_i = \frac{1}{|E_i|} \sum_{e_j \in E_i} e_j,
\end{equation}
where \( |E_i| \) is the number of EnMAP bands in \( E_i \).

The resulting set of six aggregated bands is:

\begin{equation}
    \hat{B}_{\text{EnMAP}} = \{\hat{b}_1, \hat{b}_2, \ldots, \hat{b}_6\}
\end{equation}

The $\hat{b}_i$ bands retain consistency with the HLS spectral intervals while preserving EnMAP's broader spectral coverage at a coarser resolution.



This strategy not only bridges the domain gap between HLS and EnMAP but also enhances the student model's ability to generalize across different spectral resolutions and geospatial contexts.

\subsubsection{Feature Distillation and Dimension Alignment}
\label{sec:kd_loss}

To facilitate cross-domain knowledge transfer from the teacher model (pre-trained on HLS data) to the student model (trained on EnMAP data), we employ a strategy of mid-layer feature distillation. Specifically, the student network learns from the teacher's intermediate (\textit{i.e.}, \textit{middle-layer}) feature representations, which often capture robust, domain-invariant characteristics. However, discrepancies in input channel or layer dimensionality and architectural configurations between the two models necessitate a mechanism to reconcile the resulting feature maps.


 We leverage a fully connected (FC) layer to map the teacher's mid-layer feature vectors into a dimension compatible with the student's corresponding layer. This FC layer operates as a learned transform, ensuring that feature vectors from the teacher can be aligned and compared directly to the student's hidden representations of the selected layer(s). We performed ablation studies (Section ~\ref{sec:optimal_kd_layer}) to identify the most effective teacher-student layer(s) pairings for distillation. Empirically, we find that employing features from an upper-mid encoder layer—specifically layer 8 out of 12—yields stronger knowledge distillation performance.

To facilitate the transfer of mid-layer features, we explored multiple loss functions for the teacher-student alignment, including L1 loss, Kullback–Leibler Divergence (KLD) \cite{kullback1951information}, and Jensen–Shannon (JS) Divergence \cite{journals/tit/Lin91}. Empirical results (Section ~\ref{sec:best_kd_function}) indicate that KLD outperforms the other alternatives, particularly in our HLS-to-EnMAP context, as it provides a directional measure of distributional mismatch—well-suited for bridging domain gaps and capturing subtle spectral distinctions that arise from the teacher's multispectral domain and the student's hyperspectral space.

\subsection{Spatial Features Guided Masking}

A primary objective of our HyperKD framework is to ensure that the student model (a masked autoencoder architecture) receives maximal benefit from the teacher model (the pre-trained encoder). Conventional random masking strategies often overlook intricate spatial or spectral structures crucial to effective knowledge transfer. To overcome this limitation, we introduce a feature-guided masking mechanism that uses Gabor filters \cite{gabor1946theory} and wavelet transforms \cite{meyer1992wavelets} to identify patches whose structural complexity and spectral variability are highest, thus demanding the most challenging reconstructions.



The Gabor filter, known for its ability to capture edge information and texture details \cite{mehrotra1992gabor}, is defined as:

\begin{equation}
G(x, y; \lambda, \theta, \psi, \sigma, \gamma) = \exp\left(-\frac{x'^2 + \gamma^2 y'^2}{2\sigma^2}\right) \cos\left(2\pi \frac{x'}{\lambda} + \psi\right)
\end{equation}

where $x' = x \cos\theta + y \sin\theta$ and $y' = -x \sin\theta + y \cos\theta$ represent rotated coordinates, $\lambda$ is the wavelength of the sinusoidal component, $\theta$ controls the orientation, $\psi$ is the phase offset, $\sigma$ determines the standard deviation of the Gaussian envelope, and $\gamma$ represents the spatial aspect ratio.

Complementing this, we leverage wavelet transforms to capture multi-scale spectral information. The discrete wavelet transform (DWT) decomposes an image into approximation and detail coefficients, extracting both low- and high-frequency components:

\begin{equation}
W_{\psi}(a, b) = \int_{-\infty}^{\infty} X(t) \psi^*_{a,b}(t) dt
\end{equation}

where $\psi_{a,b}(t) = \frac{1}{\sqrt{a}} \psi\left(\frac{t-b}{a}\right)$ is the scaled and translated wavelet function, and $a, b$ denote the scale and translation parameters, respectively. The response can be used individually to rank the patches or combined.


We define a patch significance score $S_p$ by applying either a Gabor filter (when focusing on edge/texture information) or a wavelet transform (when emphasizing multi-scale analysis). Specifically, depending on the chosen filtering strategy,

\begin{equation}
S_p =  \|G_p\| ~~\text{or}~~ S_p= \|W_p\|
\end{equation}

where $G_p$ and $W_p$ represent the Gabor and wavelet responses for patch $p$. For this study, we explore each method's performance separately (see more details in \autoref{tab:model_performance}).


Given an input hyperspectral image with dimension $H \times W \times C$ where $H$, $W$ and $C$ denote the height, width, and number of channels respectively, we partition the spatial domain into patches of size $p \times p$, thereby creating patches of shape $p \times p \times C$. Thus, the total number of patches becomes $N = \frac{H}{p} \times \frac{W}{p} $. 
After computing $S_p$ for each patch $p$, we rank the patches in descending order of their significance score. We then apply a masking ratio $ r \in [0,1]$ to determine the fraction of patches to be masked. Specifically, we mask the top $r \times N$ patches with the highest $S_p$ values, forcing the student autoencoder to reconstruct precisely those regions identified as the most challenging. This selective masking strategy compels the student model to reconstruct the most challenging regions, maximizing the effectiveness of knowledge transfer from the teacher model and thus enhancing the student model's ability to improve reconstruction performance.


\subsection{Features-Driven Reconstruction Loss}

Traditional pixel-wise loss functions such as Mean Absolute Error or Mean Squared Error (MSE) ensure spectral fidelity by penalizing direct deviations between reconstructed and ground-truth images. However, these metrics do not explicitly model spatial relationships or structural patterns—key aspects of hyperspectral image analysis, given the complex texture and edge information often present in such data. Moreover, purely pixel-based measures can be unduly influenced by outliers or extreme values, which may distort the overall reconstruction quality \cite{1284395,6562752}.


To address these limitations, we propose a composite loss that couples MSE with the Structural Similarity Index (SSIM) \cite{zujovic2013structural}, thereby preserving both spectral accuracy and spatial coherence.

Formally, the MSE component is defined as: 

\begin{equation}
\mathcal{L}_{\text{MSE}} = \frac{1}{N} \sum_{i=1}^{N} (X_i - \hat{X}_i)^2
\label{eq:mse}
\end{equation}

where $X_i$ and $\hat{X}_i$ denote the ground truth and reconstructed hyperspectral pixel values, respectively, and $N$ is the total number of pixels.


SSIM focuses on preserving spatial structures and perceptual quality during reconstruction \cite{zujovic2013structural}. The SSIM is formulated as:

\begin{equation}
\mathcal{L}_{\text{SSIM}} = 1 - \frac{(2\mu_X \mu_{\hat{X}} + C_1)(2\sigma_{X\hat{X}} + C_2)}{(\mu_X^2 + \mu_{\hat{X}}^2 + C_1)(\sigma_X^2 + \sigma_{\hat{X}}^2 + C_2)}
\end{equation}

where $\mu_X$ and $\mu_{\hat{X}}$ are the mean intensities of $X$ and $\hat{X}$, $\sigma_X^2$ and $\sigma_{\hat{X}}^2$ represent their variances, $\sigma_{X\hat{X}}$ is the covariance, and $C_1, C_2$ are small constants to stabilize the division.

We weight the contributions of MSE and SSIM via a linear combination:

\begin{equation}
\mathcal{L}_{\text{recon}} = \lambda_1 \mathcal{L}_{\text{MSE}} + \lambda_2 \mathcal{L}_{\text{SSIM}}
\end{equation}

where $\lambda_1$ and $\lambda_2$ are weighting factors that control the contribution of each loss term.

Finally, the feature distillation loss $\mathcal{L}_{\text{KD}}$ (detailed in Section ~\ref{sec:kd_loss}), is combined with the reconstruction loss for the effective student model training. The weighted total loss is expressed as:

\begin{equation}
\mathcal{L}_{\text{total}} = \alpha(\lambda_1 \mathcal{L}_{\text{MSE}} + \lambda_2 \mathcal{L}_{\text{SSIM}}) + \beta \mathcal{L}_{\text{KD}}
\label{eq:total_loss}
\end{equation}

where $\alpha$ and $\beta$ are hyperparameters regulating the relative influence of reconstruction quality versus knowledge transfer. By coupling pixel-level accuracy, structural fidelity, and inter-network feature alignment, this integrated loss function offers robust guidance for the student model, particularly in the context of high-dimensional hyperspectral data.

\begin{table*}
    \caption{Image reconstruction comparison: HyperKD, with specialized loss functions and masking, outperforms Student (W/O KD) and Baseline KD models on two datasets.}
    \label{tab:model_performance}
    \vskip 0in
    \begin{center}
    \setlength{\tabcolsep}{4pt}
    
  \begin{tabular}{@{}cllllll *{4}{c}|*{2}{c}@{}}
  \toprule
ID & Model & \makecell{Recon.\\Loss} & \makecell{KD\\Loss} & \makecell{Masking \\ Strategy}  & \makecell{Salient\\Patches} & \makecell{Start Rand.\\Masking} & \multicolumn{4}{c}{\textbf{Test Dataset 1}} & \multicolumn{2}{c}{\makecell{\textbf{Test Dataset 2}\\ \textbf{(Unseen Area)}}} \\
\cmidrule(lr){8-11} \cmidrule(lr){12-13}
& & & & & & & \makecell{Avg.\\ PSNR} & \makecell{Per Channel\\PSNR (Max)} & \makecell{ Avg.\\SSIM} & \makecell{Per Channel\\ SSIM (Max)} & \makecell{Avg.\\PSNR } & \makecell{Avg.\\SSIM}\\
\midrule
  
    1 & Student & HUBER & - & - & - & - & 24.61 & 32.45 & 0.55 & 0.82 & 25.48 & 0.52 \\
    2 & Base KD & HUBER  & L1 & Random  & - & epoch 1 & 27.10 & 33.49 & 0.65 & 0.85 & 27.56 & 0.62 \\
    3 & HyperKD$^\dagger$ & MSE + SSIM  & KLD & Wavelet  & visible & - & 27.70 & 34.32 & 0.70 & 0.87 & 28.05 & 0.68 \\
    4 & HyperKD$^\ddagger$  & MSE + SSIM  & KLD & Wavelet & masked & epoch 100 & 30.80 & 37.39 & 0.76 & 0.91 & 30.64 & 0.73 \\
    
    5 & \textbf{HyperKD} & MSE + SSIM & KLD & Gabor Filter & masked & epoch 100 & \textbf{31.02} & \textbf{37.56} & \textbf{0.77} & \textbf{0.91} & \textbf{30.62} & \textbf{0.73} \\
    \bottomrule
  \end{tabular}
\end{center}
\end{table*}







\begin{table}
    \caption{HyperKD's superior reconstruction across extensive geospatial regions (CA, CO, KS) highlights its advantage.}
    \label{tab:model_recon_performance_3}
    \vskip 0in
    \begin{center}
    \setlength{\tabcolsep}{4pt}

  \begin{tabular}{@{}clll *{2}{c}@{}}
  \toprule
ID & Model & \makecell{Recon.\\Loss}  & \makecell{Masking \\ Strategy}  & \multicolumn{2}{c}{ \makecell{\textbf{Test Dataset 3}\\ (CA+CO+KS)}} \\ 
\cmidrule(lr){5-6}
& & &  & \makecell{Avg.\\PSNR } & \makecell{Avg.\\SSIM}\\
\midrule
  
    1 & Student & HUBER & - & 25.57 & 0.41 \\
    2 & Base KD & HUBER  & Random   & 31.73 & 0.80 \\
    3 & \textbf{HyperKD} & MSE + SSIM & Gabor Filter & \textbf{33.62} & \textbf{0.85} \\
    \bottomrule
  \end{tabular}
\end{center}
\end{table}



\section{Experiments and Discussion}

    

\subsection{Experimental Setup}

\subsubsection{Model Configuration} We adopt the Prithvi-100M foundational model as our teacher network. Prithvi is based on a masked autoencoder framework with a vision transformer backbone, comprising 12 transformer blocks and originally designed for 18-channel inputs (6 spectral channels across 3 timestamps). Although the teacher and student share a similar ViT-based architecture, shown below, the student network accepts a significantly larger input dimension of 218 hyperspectral channels. 

\textbf{Base Model (BaseKD)} To keep the model complexity as same as Prithvi, the backbone of BaseKD is designed with the standard configuration of a small version of MAE that uses a patch size of 16 x 16 with an embedding dimensionality of 768. BaseKD contains 12 transformer layers each with 12 attention heads, and a multi-layer perceptron (MLP) hidden dimensionality of 3072. The decoder processes embeddings with a dimensionality of 512 and includes 12 Transformer layers and 16 attention heads, along with an MLP hidden dimensionality of 2048. The same architecture backbone is used for the student model and the proposed model, HyperKD even if the input dimension is significantly high. 



\subsubsection{Implementation Details} All models are trained using the Adam optimizer, paired with a custom learning rate scheduler. Input images of $224 \times 224 \times 218$ (height, width, and spectral bands) are partitioned into $16 \times 16$ patches, with a $75\%$ masking ratio applied during both training and testing. We employ a single NVIDIA A100 (80GB) GPU at an effective batch size of 32. Following the normalization strategy used by Prithvi, we standardize each spectral band to achieve zero-mean, unit-variance inputs. Inspired by ~\cite{jakubik2023foundation}, we store training and testing datasets in the Zarr format to facilitate faster I/O operations.

\subsubsection{Evaluation Metrics}

\begin{figure}[!htbp]
    \centering
    \includegraphics[width=1\linewidth]{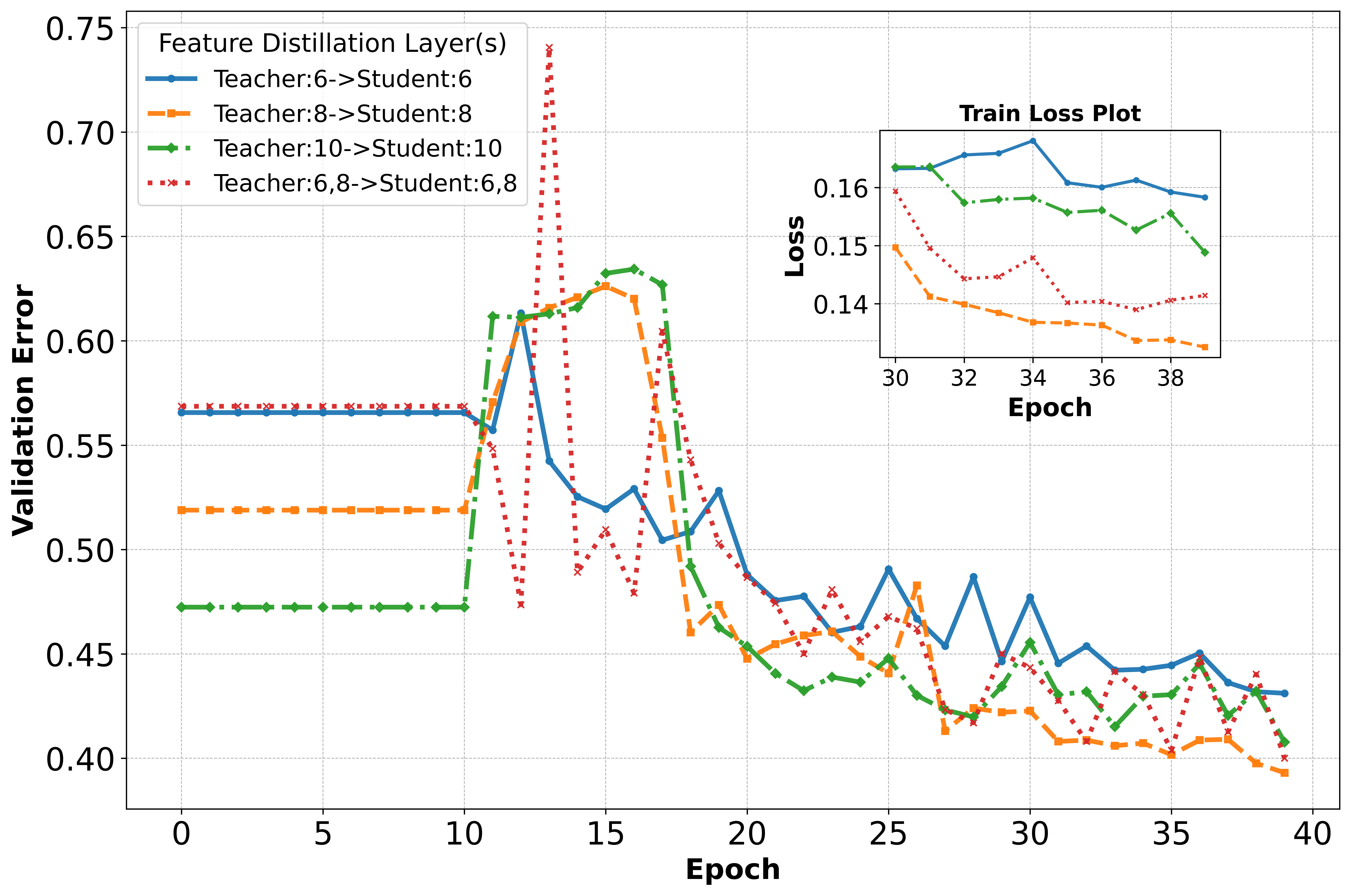}
    \caption{Impact of Teacher-Student Layer Pairings on Feature Distillation Convergence (Fig. 2). The study compares convergence trends across different layer combinations, with initial training stabilized by a 10-epoch warm-up phase.}
    \label{fig:train_valid_loss_comp}
    \vskip -0.15in
\end{figure}

To comprehensively assess HyperKD's image reconstruction accuracy, we employ the Peak Signal-to-Noise Ratio (PSNR) and the Structural Similarity Index (SSIM). PSNR captures pixel-level fidelity, making it particularly relevant for evaluating spectral and spatial accuracy in geospatial contexts. SSIM quantifies the structural and textural similarities between the reconstructed and ground-truth images, thereby providing insights into the preservation of spatial patterns.

\begin{figure*}[!htbp]
    \centering
    \includegraphics[width=1\linewidth]{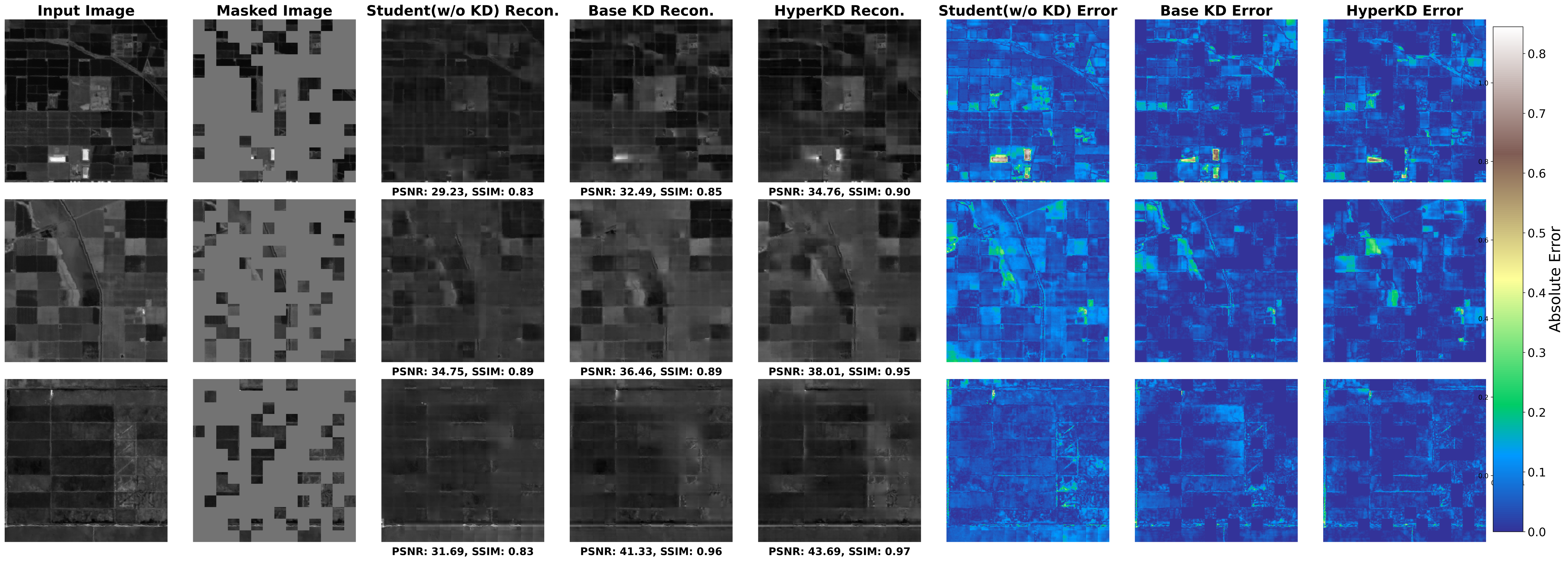}
    \caption{Image Reconstruction Quality Comparison: HyperKD vs. Baseline KD vs. Student Model Without KD. HyperKD achieves superior performance across all test samples, demonstrating the highest PSNR and SSIM scores in the 218-band spectral analysis.}
    \label{fig:recons_img_compl}
\end{figure*}

\subsection{Ablation Experiments}
\label{sec:ablation_experiments}

Given the high computational demands of training our model, we conducted several ablation studies on a reduced dataset consisting of 1000 training samples, 100 validation samples, and 300 test samples. We also limited the number of epochs (40 for knowledge distillation [KD] loss functions, and 150 for KD layers) to streamline experimentation. All experiments were performed with a 75\% masking ratio and MSE for reconstruction loss, enabling a controlled comparison of different KD configurations. 



\subsubsection{Selecting the Optimal KD Layer}
\label{sec:optimal_kd_layer}

Both the teacher and student networks share a Vision Transformer-based Masked Autoencoder (ViT-MAE) backbone comprising 12 transformer blocks. Since lower layers primarily capture low-level features, we performed an ablation study to determine the most informative feature representations within the mid-to-top layers. As shown in \autoref{fig:train_valid_loss_comp}, feature distillation at layer 8 yielded better student model convergence compared to other layers.
Therefore, we selected \textbf{layer 8} as the KD layer for all subsequent experiments.



\begin{table}
    \caption{Ablation: KD Function Impact on Reconstruction Accuracy with the BaseKD model (\autoref{tab:model_performance}).}
    \label{tab:kd_functions}
    \vskip -0in
    \centering
    \begin{tabular}{l c c c c}
        \toprule
        \makecell{KD Loss \\  Funciton} & \makecell{ Average \\ PSNR} & \makecell{Per Channel \\ PSNR (Max)} & \makecell{ Average \\ SSIM} & \makecell{Per Channel \\ SSIM (Max)} \\
        \midrule
        L1  & 26.86 & 32.78 & 0.63 & 0.84 \\
        KLD & 27.39 & 33.48 & 0.65 & 0.85 \\
        JS  & 27.08 & 33.36 & 0.64 & 0.85 \\
        \bottomrule
    \end{tabular}
    \vskip -0.15in
\end{table}

\subsubsection{Choosing the Best KD Function}
\label{sec:best_kd_function}

To evaluate the influence of different KD loss functions, we compared L1, Kullback–Leibler Divergence (KLD), and Jensen–Shannon (JS) Divergence, reporting both PSNR and SSIM metrics. As summarized in \autoref{tab:kd_functions}, \textbf{KLD} provided the highest PSNR (27.39) and SSIM (0.65), reflecting better structural preservation. In contrast, L1 performed moderately (PSNR: 26.86, SSIM: 0.63), while JS divergence resulted in lower scores (PSNR: 27.08, SSIM: 27.08), indicating diminished effectiveness for our training setup. Based on these findings, we selected \textbf{KLD} as the principal KD function for its superior feature-preservation capabilities.




\subsection{Effectiveness of the Specialized Loss Function}

 Our proposed composite loss function (\autoref{eq:total_loss}) substantially enhances reconstruction performance by preserving both spatial coherence and pixel-level fidelity. As reported in \autoref{tab:model_performance}, incorporating the Structural Similarity Index (SSIM) increases average SSIM from 0.65 to 0.77, while simultaneously improving the average Peak Signal-to-Noise Ratio (PSNR) from 27.10 to 31.02. This increase underscores the capacity of the student model to maintain fine-grained spatial details during reconstruction. Furthermore, using Kullback–Leibler Divergence (KLD) in the knowledge distillation process refines the student model's feature space alignment with that of the teacher, leading to an even closer approximation of the teacher's output.


\subsection{Impact of the Spatial Features Driven Masking Strategy}

Adopting advanced masking strategies based on Gabor filters or wavelet transforms notably improves reconstruction. As shown in \autoref{tab:model_performance}, wavelet-guided masking combined with our specialized loss function elevates the average PSNR and SSIM to 30.80 and 0.76, respectively—exceeding the values obtained using random masking (27.10 PSNR, 0.65 SSIM). Similarly, a Gabor filter-based masking approach also achieves strong performance, with the highest observed PSNR of 31.02 arising from an initial application of Gabor masking followed by a random masking finetuning stage. From \autoref{fig:recons_img_compl}, HyperKD outperforms in both standard reconstruction and capturing high spatial variability, even for unseen regions (Test Set 2, \autoref{tab:model_performance}).


\subsection{Performance of HyperKD for Inverse-Shift Spectral Domain Adaptation}

The results summarized in \autoref{tab:model_performance} highlight the efficacy of the HyperKD framework in mitigating the inverse domain shift, wherein the student model is trained on 218 hyperspectral channels while receiving knowledge from a teacher (Prithvi) equipped with only 6 synthetic hyperspectral channels. Despite this substantial gap in dimensionality, the proposed knowledge distillation approach substantially improves reconstruction, achieving an average PSNR of 31.02 and an average SSIM of 0.77. These findings illustrate the ability of HyperKD to transfer robust representations from lower-dimensional to higher-dimensional spectral domains, thereby preserving both spectral fidelity and structural detail. The consistently superior reconstruction accuracy of HyperKD, as demonstrated in \autoref{tab:model_recon_performance_3} across the expansive geospatial regions of CA, CO, and KS, strongly corroborates the effectiveness of our proposed methodology.


\subsection{Spectral Channel Level Reconstruction Performance of HyperKD}

\begin{figure}[!htbp]
\vskip -0.1in
  \centering
    \includegraphics[ width=1\linewidth, alt={Bandwise PSNR}]{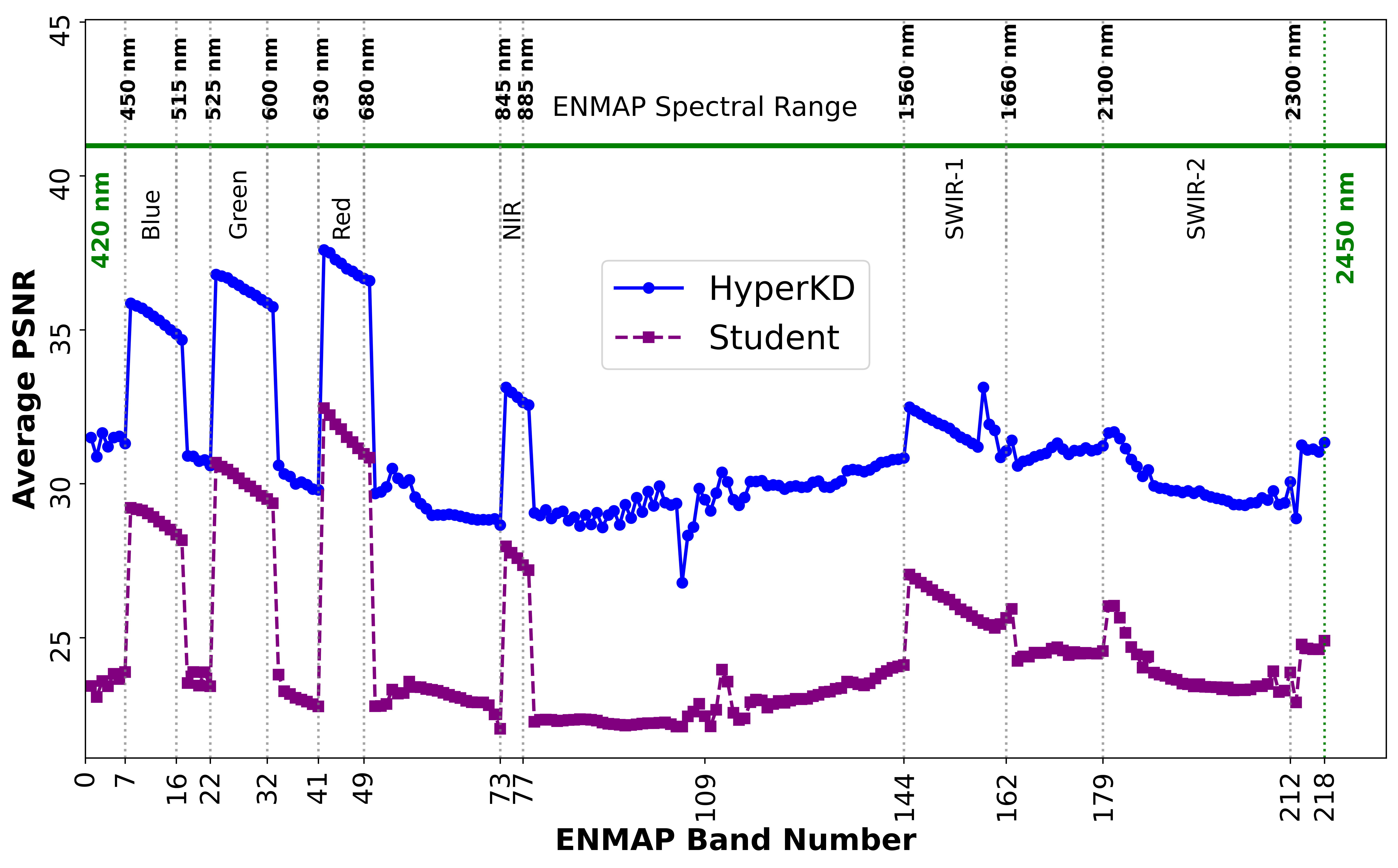}
  \caption{Comparison of reconstruction accuracy at individual channel levels.}
  \label{fig:bandwise_psnr}
  \vskip -0.1in
\end{figure}

Figure~\ref{fig:bandwise_psnr} illustrates the spectral overlaps between HLS and EnMAP, along with each dataset's band numbering. This mapping was utilized to synthesize 6 spectral bands, as described in Section~\ref{sec:hls_enmap_alignment}—to facilitate knowledge transfer from the HLS-based teacher model to the EnMAP-based student.

The experimental results demonstrate that the HyperKD framework achieves substantially higher reconstruction accuracy across all EnMAP bands compared to the student model trained without KD. Notably, the improvements are most pronounced in regions where HLS and EnMAP bands overlap, indicating that the teacher model, which was originally trained on HLS data, provides effective guidance in this shared spectral range. In non-overlapping regions, performance gains are somewhat smaller due to a lack of direct spectral correspondence. 

Overall, these findings underscore the ability of HyperKD to address the inverse domain gap challenge by transitioning from a teacher model with fewer spectral channels (HLS) to a student model equipped with a broader EnMAP spectrum. Beyond enhancing accuracy in overlapping bands, the teacher's representation further supports generalized performance throughout the entire EnMAP spectral range, highlighting the potential of our approach to strengthen cross-domain transferability in hyperspectral image reconstruction.




\section{Downstream Tasks \& Performance}

The pretrained \textbf{HyperKD} model offers significant potential for a range of downstream tasks that involve hyperspectral imagery (HSI) input data. This versatile model can be adapted for both regression and classification tasks by selecting the appropriate downstream head. In this paper, we specifically explore two classification tasks of land cover classification and crop type identification, and one regression task of soil organic carbon (SOC) prediction using the same pretrained HyperKD encoder. Detailed about the data and the modeling approach are provided in the following section.


\begin{figure}[!htbp]
  \centering
    \includegraphics[width=1\linewidth, alt={Downstream Model}]{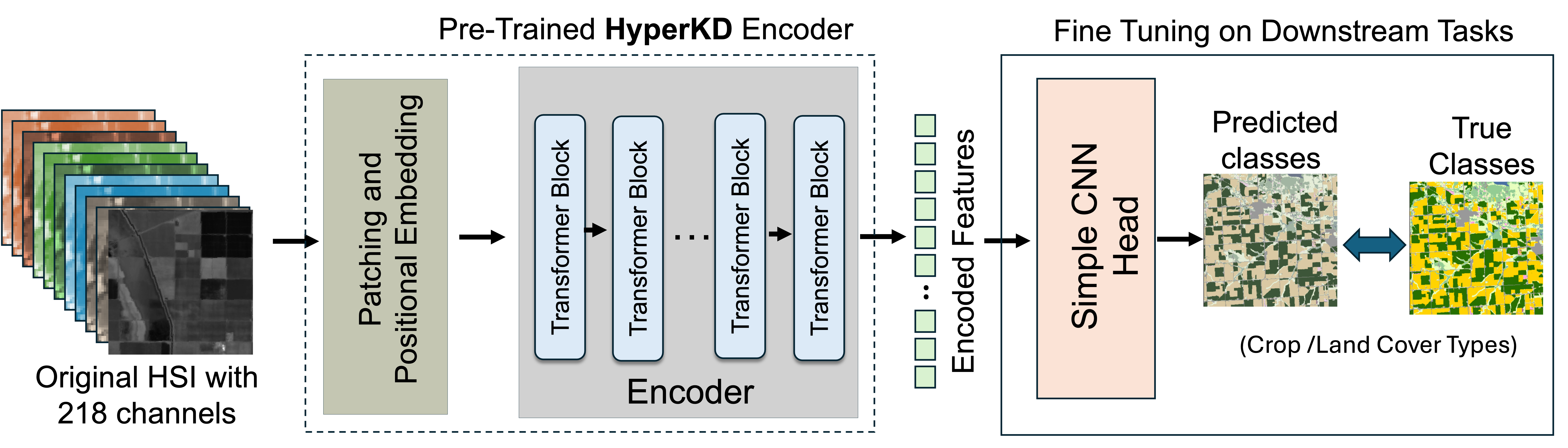}
  \caption{Downstream task implementation consisting of the frozen HyperKD encoder (pre-trained via knowledge distillation) and a CNN head.}
  \label{fig:downstream_architecture}
\end{figure}
In this work, we concentrate on two classification scenarios—\textit{land cover classification} and \textit{crop type identification} using EnMAP hyperspectral data.

\subsubsection{Downstream Model}

As illustrated in \autoref{fig:downstream_architecture}, we employ the pre-trained HyperKD encoder, which processes 218 EnMAP spectral bands and appends a lightweight classification head consisting of a small number of convolutional layers. The pre-trained encoder remains frozen to preserve learned spectral-spatial features, while the classification head is fine-tuned onthe  target dataset. This setup allows us to leverage the robust representations acquired from knowledge distillation without incurring the computational expense of retraining the entire network. Similarly, a simple CNN regression head is plugged in with the pretrained encoder for the regression task.

\subsubsection{Downstream Task Evaluation Metrics}

To assess classification performance, we adopt two primary metrics: (i) Top-1 accuracy, which measures the proportion of correctly classified samples, and (ii) Mean Intersection over Union (mIoU), a popular metric that quantifies the overlap between predicted and true labels.

\subsection{Land Cover Classification} \label{sec:nlcd}

For land cover classification, we utilize the 30m resolution National Land Cover Database (NLCD) \cite{us2025annual}, which categorizes broad land cover classes. Given that NLCD is not updated frequently, we align its timestamps with the closest available EnMAP acquisitions to maximize temporal consistency. We focus on five land cover categories in California that can be robustly paired with EnMAP observations: \textit{Shrub/Scrub}, \textit{Herbaceous}, \textit{Cultivated Crops}, \textit{Forest}, and \textit{Other}.

\begin{table}
\caption{Land Cover Classification on NLCD data ( top 5 classes in CA) }
\label{tab:nlcd_performance}
\vskip 0in
\begin{center}
\resizebox{\columnwidth}{!}{ 
\begin{tabular}{l c c c c c}
\toprule
\textbf{Class Name} & \multicolumn{2}{c}{\textbf{Student + CNN Head}} & \multicolumn{2}{c}{\textbf{HyperKD + CNN Head}} \\
\cmidrule(lr){2-3} \cmidrule(lr){4-5}
 & \textbf{Top-1 Acc (\%)} & \textbf{mIoU (\%)} & \textbf{Top-1 Acc (\%)} & \textbf{mIoU (\%)} \\
\midrule
Shrub/Scrub & 58.41 & 41.90 & \textbf{59.38} & \textbf{41.85} \\
Herbaceous & 61.06 & 41.00 & 58.07 & 38.39 \\
Cultivated Crops & 90.80 & 75.29 & \textbf{92.91} & \textbf{75.67} \\
Forest & 84.67 & 57.76 & 83.21 & \textbf{58.50} \\
Other & 71.67 & 61.36 & \textbf{72.78} & \textbf{63.70} \\
\midrule
\textbf{Mean} & 75.42 & 55.46 & \textbf{75.49} & \textbf{55.62} \\
\bottomrule
\end{tabular}
}
\end{center}
\vskip -0.1in
\end{table}

\begin{table}[!htbp]
\caption{Land Cover Classification Accuracy on NLCD (CA+CO+KS)"}
\label{tab:nlcd_performance_2}
\vskip 0in
\begin{center}
\begin{tabular}{l c c c}
\toprule
\textbf{Class Name} & \multicolumn{3}{c}{Top-1 Accuracy (\%)} \\
\cmidrule(lr){2-4}
& \textbf{Student} & \textbf{BaseKD} & \textbf{HyperKD} \\
\midrule
Shrub/Scrub         & 63.18 & 62.00 & \textbf{63.56} \\
Herbaceous          & 35.43 & 38.70 & \textbf{46.25} \\
Cultivated Crops    & 56.70 & 77.18 & \textbf{85.50} \\
\bottomrule
\end{tabular}
\end{center}
\vskip 0in
\end{table}

\autoref{tab:nlcd_performance} demonstrates that the proposed knowledge distillation (KD) approach, incorporating Gabor filter-based masking, outperforms a standalone student model for the most of classes. While improvements differ among classes, substantial gains are observed in the Cultivated Crops and Other categories, suggesting that KD effectively captures complex spatial textures. Despite inherent class imbalances and occasional noise in the NLCD labels—particularly in California’s diverse landscapes—our single-timestamp EnMAP approach compares favorably to the Prithvi model, which relies on three input timestamps. These findings underscore the efficiency of HyperKD in learning discriminative land cover representations under constrained temporal conditions. The consistently enhanced downstream performance of HyperKD across the expanded regions of CA, CO, and KS, as detailed in \autoref{tab:nlcd_performance_2}, underscores the significance of pre-training the foundational model using our proposed approach.

\subsection{Crop Type Identification} \label{sec:cdl}

We evaluate the Cropland Data Layer (CDL) \cite{copenhaver2021examining}, another 30m resolution product that focuses on distinguishing specific crop types. Similar to NLCD, we choose the closest EnMAP timestamps available for accurate spatial alignment. For this study, we concentrate on four major crop categories in California—\textit{Shrubland}, \textit{Grassland/Pasture}, \textit{Forest}, and \textit{Other} aligning with our available EnMAP data.

\begin{table}
\caption{Crop Type Identification on CDL data ( top 4 classes in CA) }
\label{tab:cdl_performance}
\vskip 0in
\begin{center}
\resizebox{\columnwidth}{!}{ 
\begin{tabular}{l c c c c}
\toprule
\textbf{Class Name} & \multicolumn{2}{c}{\textbf{Student + CNN Head}} & \multicolumn{2}{c}{\textbf{HyperKD + CNN Head}} \\
\cmidrule(lr){2-3} \cmidrule(lr){4-5}
 & \textbf{Top-1 Acc (\%)} & \textbf{mIoU (\%)} & \textbf{Top-1 Acc (\%)} & \textbf{mIoU (\%)} \\

\midrule
Shrubland           & 40.31 & 34.74 & \textbf{55.93} & \textbf{46.00} \\
Grassland/Pasture   & 22.74 & 16.85 & \textbf{25.95} & \textbf{19.50} \\
Evergreen Forest    & 81.72 & 53.04 & \textbf{93.62} & \textbf{56.44} \\
Other               & 90.87 & 62.02 & 87.68 & 72.44 \\
\midrule
\textbf{Mean}       & 67.41 & 41.66 & \textbf{73.27} & \textbf{48.60} \\
\bottomrule
\end{tabular}
}
\end{center}
 \vskip -0.1in

\end{table}

As visible in \autoref{tab:cdl_performance}, HyperKD generally surpasses the baseline student across key crop categories. While the magnitude of improvements varies, the KD approach yields notable benefits in structurally intricate classes, reinforcing its capacity to harness teacher-model features that are critical for capturing subtle inter-class boundaries. Although CDL is known to contain noisy labels and uneven class distributions in certain parts of California, our single-timestamp strategy remains competitive with three-timestamp baselines (\textit{e.g.}, Prithvi), highlighting the adaptability and robustness of HyperKD representations.


NLCD and CDL datasets are inherently noisy, and for the aligned timeframe and region with ENMAP, the dataset is imbalanced. Despite these challenges, our model achieves strong performance compared to the student-only and traditional baseline KD models relying on a single timestamp input. This highlights the efficiency of our model in learning meaningful representations with limited temporal data.


\begin{table}[!htbp]
\caption{Soil Organic Carbon Prediction Accuracy}
\label{tab:soc_prediction}
\vskip 0in
\centering
\begin{tabular}{l c}
\toprule
\textbf{Model} & \textbf{SOC Prediction (MAE)} \\
\midrule
Student w/o KD  & 0.045 \\
Baseline KD   & 0.046 \\
HyperKD w Gabor Filter  & \textbf{0.038} \\
\bottomrule
\end{tabular}
\end{table}

\subsection{Soil Organic Carbon Prediction} \label{soc:prediciton}
To evaluate the efficacy of our approach for a regression task, we focused on predicting soil organic carbon (SOC) content. Our target variable consisted of the top 30-centimeter SOC values extracted from the 2023 30-meter gNATSGO dataset \cite{soil2019gridded}. We spatially aligned these SOC data points with our pre-processed ENMAP hyperspectral tiles covering the diverse agricultural landscapes of California, Colorado, and Kansas, thereby creating a geographically representative train and test split. To ensure stable model training, we applied a standard mean-standard deviation normalization to the SOC values. The results, shown in \autoref{tab:soc_prediction}, clearly indicate that the HyperKD pre-trained encoder significantly enhances the accuracy of downstream SOC predictions, outperforming both the student model trained from scratch and a standard knowledge distillation baseline. This outcome logically implies that our proposed cross-modal knowledge transfer strategy, which effectively bridges the spectral dimensionality gap between the foundational and target models, yields robust and generalizable feature representations that are beneficial not only for spectral classification but also for continuous value regression tasks in remote sensing applications.

\section{Conclusion \& Future Work}
In this work, we introduced \textbf{HyperKD}, a novel knowledge distillation framework designed to address the challenges of inverse domain adaptation in hyperspectral imaging (HSI). HyperKD enables robust and reliable representation learning from hyperspectral satellite imagery, which is characterized by extremely low temporal frquency and high spectral dimensionality. The framework effectively transfers spectral relationships and their geospatial patterns from a foundational model (Prithvi-100M), which is trained on multispectral data with lower spectral resolution but significantly higher temporal frequency and broader geospatial coverage. HyperKD successfully bridged the spectral gap and achieved significant improvements in both reconstruction and downstream task performance. Our advanced masking strategies (Gabor filters and wavelet transforms) with a multi-component loss function (MSE, SSIM, and KLD feature distillation) demonstrated superior reconstruction accuracy. Furthermore, HyperKD achieved competitive results in downstream tasks, with strong Top-1 Accuracy and mIoU scores, demonstrating the effectiveness of our methodology in mitigating the inverse domain shift.


Regarding future research directions, we plan to extend the proposed knowledge distillation approach to a broader range of geospatial conditions, including diverse climatic regions and varying spatial extents. In the current study, both the teacher and student models were trained on satellite imagery with the same spatial resolution. In our future research, we aim to adapt our methodology to handle remote sensing datasets with varying spatial resolutions. Finally, we will explore the applicability of our approach to a wider range of downstream tasks and geospatial applications.


\bibliographystyle{IEEEtran}
\bibliography{biblio}

\end{document}